%% file: main.tex
\documentclass{article}

\usepackage[nonatbib, preprint]{neurips_2022}

\usepackage[utf8]{inputenc} % allow utf-8 input
\usepackage[T1]{fontenc}    % use 8-bit T1 fonts
\usepackage{hyperref} % hyperlinks
\usepackage{url}            % simple URL typesetting
\usepackage{booktabs}       % professional-quality tables
\usepackage{amsfonts}       % blackboard math symbols
\usepackage{nicefrac}       % compact symbols for 1/2, etc.
\usepackage{microtype}      % microtypography
\usepackage{xcolor}         % colors
\usepackage{amsmath}
\usepackage{pdfpages}
\usepackage[colorinlistoftodos]{todonotes}
\usepackage{subcaption}
\usepackage{algorithm} 
\usepackage{algpseudocode} 
\usepackage{mathtools}   
\usepackage{verbatim}

\title{XOOD: Extreme Value Based Out-Of-Distribution Detection For Image Classification}

\let\[\equation
\let\]\endequation

\author{%
  Frej Berglind\\
  Louisiana State University\\
  Baton Rouge, LA 70803\\
  \texttt{fbergl1@lsu.edu} \\
 \And
 Haron Temam\\
  Louisiana State University\\
  Baton Rouge, LA 70803\\
  \And
 Supratik Mukhopadhyay\\
  Louisiana State University\\
  Baton Rouge, LA 70803\\
  \And
  Kamalika Das\\
  Intuit Inc.\\
  2700 Coast Ave \\
Mountain View CA 94043\\
  \And
  Md Saiful Islam Sajol\\
  Louisiana State University\\
  Baton Rouge, LA 70803\\
  \And 
  Sricharan Kumar\\
   Intuit Inc.\\
  2700 Coast Ave \\
Mountain View CA 94043\\
  \And
  Kumar Kallurupalli\\
  Intuit Inc.\\
  2700 Coast Ave \\
Mountain View CA 94043\\
}

\begin{document}

\maketitle

\begin{abstract}
Detecting out-of-distribution (OOD) data at inference time is crucial for many applications of machine learning. We present  \texttt{XOOD}: a novel extreme value-based OOD detection framework for image classification that consists of two algorithms. The first, \texttt{XOOD-M}, is completely  unsupervised, while the second \texttt{XOOD-L} is self-supervised. Both algorithms rely on the signals captured by the extreme values of the data in the activation layers of the neural network in order to distinguish between in-distribution and OOD instances. We show experimentally that both \texttt{XOOD-M} and \texttt{XOOD-L} outperform state-of-the-art OOD detection methods on many benchmark data sets in both efficiency and accuracy, reducing false-positive rate (FPR95) by 50\%, while improving the inferencing time by an order of magnitude.

\end{abstract}

\section{Introduction}
Deep neural networks are known to be opaque in their decision-making process \cite{jeyakumar2020can}. This becomes problematic when decisions need to be made on inputs whose salient characteristics are different than what the neural network has been trained to identify. Such situations frequently arise when dealing with out-of-distribution (OOD) data \cite{yang2021generalized}, i.e., test data at inference time that does not come from the training distribution. For example, a neural network that has been trained to classify images of horses and giraffes, when provided with an image of an elephant, will classify it either as a horse or a giraffe, rather than determining that it does not belong to either of the classes.   Indeed, the fundamental assumption in supervised machine learning  that both the training and the test data come from the same probability distribution \cite{bishop1995neural,bishop2006pattern}, is violated in many real-world applications like text extraction from documents, medical diagnosis, autonomous driving, etc. In such situations, neural networks can make erroneous decisions rather than issuing a warning that they have encountered OOD data on which their decisions cannot be trusted.   When document text extraction models are deployed in real life applications, they may encounter a crumpled document that is damaged beyond recognition. Instead of informing the users of the situation, the model will attempt to extract a (possibly incorrect) set of texts from it. In safety critical decision making such as medicine, finance, and autonomous driving, for machine learning models to be trusted, it is important for the model to identify when to abstain or require human intervention \cite{gunning2019xai}. 

It is well known \cite{baseline,clune} that the ``class probabilities" output by the softmax layer of a neural network are only weakly correlated with  how confident the model should be about the prediction, even in the relatively simple case of distinguishing Gaussian noise from in-distribution data \cite{baseline}. Measuring uncertainty associated with the decisions of a neural network during inference time is an active area of research \cite{ovadia2019can}. For uncertainties stemming from encountering OOD data at test time, various supervised and unsupervised methods have been proposed in the literature \cite{sastry2020detecting}, \cite{odin}, \cite{devries2018learning}, \cite{sun2021react} that provide varying degree of accuracy and come with varying degrees of computational overhead. Since convolutional layers have linear complexity, OOD detection algorithms for convolutional networks should ideally have  linear complexity in order to be useful and scalable in real life applications. In this paper, we propose  a class of extreme value based OOD detection algorithms that are linear in complexity with respect to the size of the embedding space, and are, therefore, highly efficient and scalable, while matching or beating state of the art results in accuracy metrics. 

\begin{figure}
    \centering
    \begin{subfigure}[b]{0.3\textwidth}
         \centering
         \includegraphics[width=\textwidth]{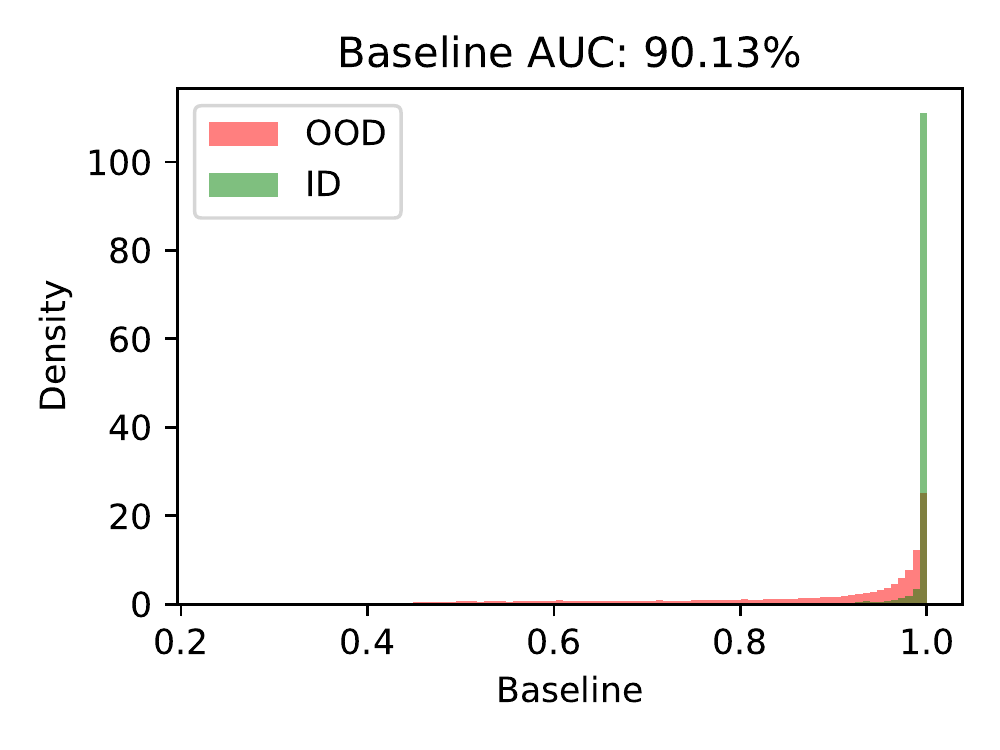}
         \caption{Baseline}
         \label{Baseline}
    \end{subfigure}
     \begin{subfigure}[b]{0.3\textwidth}
         \centering
         \includegraphics[width=\textwidth]{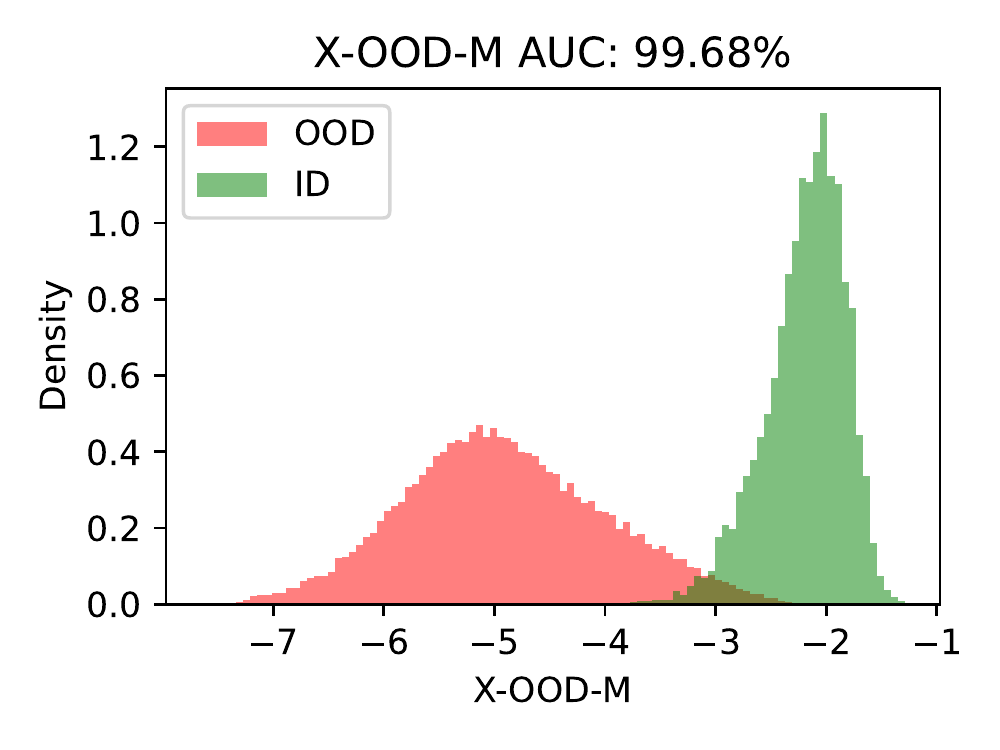}
         \caption{XOOD-M}
         \label{XOOD-M}
    \end{subfigure}
    \begin{subfigure}[b]{0.3\textwidth}
         \centering
         \includegraphics[width=\textwidth]{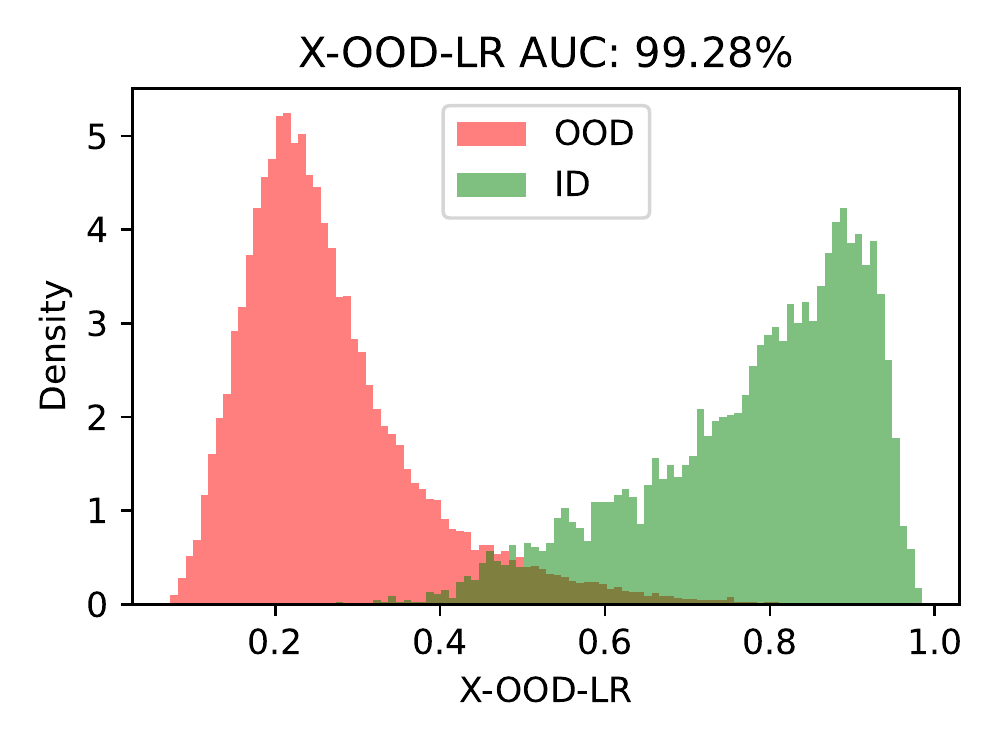}
         \caption{XOOD-L}
         \label{XOOD-LR}
    \end{subfigure}
    \caption{Comparison of the baseline result from the softmax output of a ResNet34 to our out-of-distribution detection algorithms. In this example, cifar-10 is in-distribution (ID) and SVHN is out-of-distribution (OOD). Note how confident the baseline is even though the images in SVHN do not belong to any of the classes in cifar-10.}
\end{figure}

Our \texttt{XOOD} framework (Extreme Value-based OOD Detection) that comprises two efficient and accurate algorithms for OOD detection \emph{for image classification}: the first, \texttt{XOOD-M},  being completely  unsupervised (see Section \ref{md}), while the second \texttt{XOOD-L} being self-supervised (see Section \ref{lr}). Neither of the two algorithms require prior knowledge of the OOD data, but instead rely on the computation of global extrema of the input features. Figure \ref{Baseline} presents the softmax output of a ResNet34 with CIFAR-10 as in-distribution and SVHN as OOD. As one can see, the softmax output is unable to separate between the in-distribution and the OOD images. Figures \ref{XOOD-M} and \ref{XOOD-LR} respectively show that both the unsupervised as well as the self-supervised algorithms of \texttt{XOOD} (\texttt{XOOD-M} and \texttt{XOOD-L}) are able to separate the in-distribution from the OOD images. In addition, both algorithms are robust to both uniform and Gaussian noise (see Table \ref{reacttable} in Appendix A).

\begin{comment}
Both algorithms \texttt{XOOD-M} and \texttt{XOOD-L} outperform \cite{sastry2020detecting} and \cite{lee2018simple} in terms of inference time by an order of magnitude (see Tables \ref{table3} and  \ref{table4}). In addition, both algorithms are robust to both uniform and Gaussian noise (see Table \ref{reacttable} in Appendix A). 
 On most benchmark in-distribution-OOD dataset combinations, both \texttt{XOOD-M} and \texttt{XOOD-L} outperform state-of-the-art algorithms for standard architectures \cite{baseline,odin, lee2018simple} \cite{odin,sastry2020detecting,lee2018simple,sastry2020detecting} (see Tables \ref{Table:DenseNet} and \ref{Table:ResNet}). 
 \end{comment}

%% We should mention that many OOD-algorithms such as Gram and Mahalanobs have high complexity. Since convolutional layers have linear complexity, OOD algorithms for convolutional networks should at most have linear complexity. Otherwise they won's scale well.

\paragraph{Contributions:} This paper makes the following contributions
\begin{itemize}
\item Based on the observation that the extreme value distribution of the feature space in any activation layer is unimodal on the in-distribution data, we propose a highly effective method for OOD detection using these extrema.
\item We introduce a regularized Mahalanobis distance on the extracted extreme values for unsupervised OOD detection (\texttt{XOOD-M}).
\item We present a self-supervised method using logistic regression on the extracted extreme values for OOD detection (\texttt{XOOD-L}).
\item We show experimentally that both \texttt{XOOD-M} and \texttt{XOOD-L} outperform state-of-the-art OOD detection methods on many benchmark data sets across several detection metrics. In fact, it improves the average false positive rate at 95\% true positive rate by as much as 50\% compared to the best performing existing approach \cite{sastry2020detecting}.
\item Both algorithms have an order-of-magnitude lower computational overhead compared to several other state-of-the-art OOD detection methods and scale linearly with the size of the embedding space, making them ideal for many real-life applications.
\end{itemize}

\section{Proposed Approach}
In this section, we formally define the problem statement, and discuss the theoretical background of our proposed methods before describing the two algorithms for OOD detection.

\def\t{\tilde}
\subsection{Problem Formulation}
A neural network based image classifier is a function  $f_\theta:X\rightarrow Y $, where $\theta$ represents the weights of the network,  $X$ is the input domain defined by a set of images and $Y=\{0,1,\ldots, K-1\}$, where $K$ is the number of classes in the input distribution. In the traditional image classification setting, the trained model $f_\theta$ outputs a decision $\hat{y} \in Y$ for every test image $\t{X}$. However, if $\t{X}$ is out of distribution with respect to the input distribution of $X$, we would want the model $f_\theta$ to return, along with the class label $\hat{y}$, a score akin to model confidence that can indicate whether $\t{X}$ is out of distribution with respect to the marginal probability distribution $p_{X}$ for the joint distribution $p(X,Y)$. In other words, we want to learn a function $g$ that satisfies the following condition
\[
  g(\t{X}) =
  \begin{cases*}
                                   1 & \text{if $\t{X} \sim p_{X}$} \\ 
                                   0 & \text{otherwise}
  \end{cases*}.
\]

%%where $\tilde{X}$ is the input domain, and for simplicity, we assume that the output domain $Y=\{0,1,\ldots, C-1\}$, where $C$ is the number of classes. 

%%We assume an unknown (joint) probability distribution $p(\tilde{X},Y)$ over $\tilde{X}\times Y$. Let $p_{\t{X}}$ be the marginal distribution of $\t{X}$ for the joint distribution $p$.  

%%Let the training data be  $T=\{(\t{x_1},\t{y_1}),\ldots,(\t{x_N},\t{y_N})\}$ where $\{\t{x_1},\ldots,\t{x_N}\} \subset \t{X}$ and $\t{y_i} \in Y$.   

%%We want a model $\t{f_\theta}:\t{X}\rightarrow Y\times \{0,1\}$ that, after training on $T$, will, for a test data point $\t{x} \in \t{X}$, output a decision $y\in Y$ along with a confidence $c \in \{0,1\}$.  Thus, $\t{f_\theta}(\t{x})[0]\in Y$ while $\t{f_\theta}(\t{x})[1] \in \{0,1\}$. We want the network $\t{f_\theta}$ to satisfy the following condition.
%%\[
%%  \t{f_\theta}(\t(x))[1] =
%%  \begin{cases*}
%%                                   1 & \text{if $\t{x} \sim p_{\t{X}}$} \\
%%                                   0 & \text{otherwise}
%%  \end{cases*}.
%%\]
% In practice, we want $\t{f_\theta}$ to be such that $\t{f_\theta}[1]$ satisfies the following: $\forall \t{x},\t{x'} \in \t{X}\: p_{\t{X}}(\t{x})\geq p_{\t{X}}(\t{x'}) \\Longrightarrow $

\subsection{Extreme Value Based Out-Of-Distribution Detection}
We propose two new extreme value based out-of-distribution detection algorithms that rely on the signals captured by the extreme values of the data in the activation layers of a neural network to distinguish between in distribution and OOD instances. Neither of the two algorithms require prior knowledge of the OOD data. The first algorithm, called \texttt{XOOD-M}, is completely unsupervised. It only computes some statistics on the extreme value distribution of the training data and uses the notion of  distance from those statistics to identify OOD data points. This method is extremely efficient and accurate and produces state-of-the-art results on all benchmark data sets for which the extreme value distribution of the training data is compact and well separated from the OOD data. For situations where the OOD data may not be significantly different than the in-distribution data, we propose a modified version of the \texttt{XOOD-M} algorithm called \texttt{XOOD-L}. This is a self-supervised algorithm in that it does not need to be trained on labeled OOD detection data. Instead, it learns a logistic regression model on the extreme values of the in-distribution data along with a range of distortions of that data as a surrogate for the OOD data. \texttt{XOOD-L} also beats state-of-the-art OOD detection methods on many benchmark data sets in both efficiency and accuracy.

We describe the details of both the algorithms in the next few sections. 

% The first (XOOD-L) is self-supervised and uses logistic regression trained on a range of distortions of a test dataset.  The second algorithm is completely unsupervised and uses regularized Mahalanobis distance to characterize OOD data points. Both begin by extracting extreme values before each activation layer of the model. We then use these extreme values for detecting outliers through self-supervised logistic regression or unsupervised Mahalanobis distance.

\begin{figure}[h]
    \centering
    \includegraphics[width=.4\textwidth]{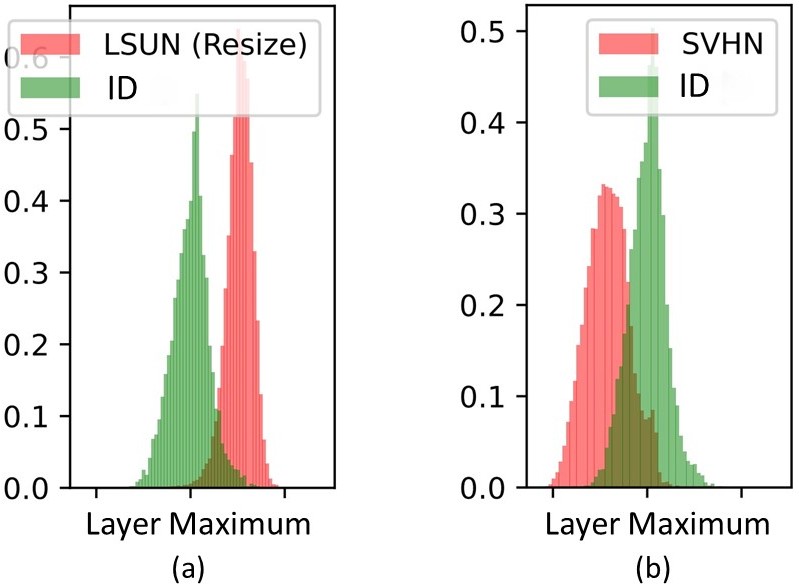}
    \caption{Distribution of maximum values before an activation layer inside ResNet34. Green represents CIFAR-10 (ID) and  Red represents LSUN-Resize(OOD) (left figure) or SVHN(OOD) (right figure). Note how in-distribution data lies in a single mode, while out-of-distribution data can be on either side. Complete examples of extreme values can be found in Appendix C}
    \label{max_example}
\end{figure}

\subsubsection{Extreme Value Extraction}
The essence of both the algorithms is in the nature of the distribution of the extreme values of the data as it enters the activation layers in a neural network. We observe that the per activation layer distribution of maximum or minimum value is consistently unimodal on in-distribution data, as can be seen in Figure \ref{max_example} and Appendix C. Based on this observation, we compute the global maximum and minimum of the input images before each activation layer in the architecture. To improve the symmetry of this distribution, we apply a Yeo-Johnson transform \cite{yeo-johnson} fitted on the extreme values of the training data. The output of this transform is standardized to zero mean and unit variance. This computation does not make use of class labels or image topology, yet it is highly effective for OOD detection.

\begin{algorithm}
\caption{\textbf{Function:} ExtremeValueExtraction}\label{fun:fe}
\begin{algorithmic}
\Require Data set $X_t$
\For{\texttt{each image $X_{t,i}$ in training set} $X_t$}
\State $X_{min,i}$, $X_{max,i}$ = \texttt{MinMax}($X_{t,i}$)
\EndFor
\State $X_{min}$ = $[X_{min,1}, X_{min,2}, X_{min,3}, X_{min,4},\dots ]$
\State $X_{max}$ = $[X_{max,1}, X_{max,2}, X_{max,3}, X_{max,4},\dots ]$

\State \textbf{return} $X_{min}$, $X_{max}$
\end{algorithmic}
\end{algorithm}

\subsubsection{Unsupervised Out-of-Distribution Detection: \texttt{XOOD-M}}\label{md}

Once the extreme values are computed for each activation layer of the neural network, a regularized Mahalanobis distance can be used for OOD detection. Let $\mu$ and $M$ be the mean and covariance of the extreme values on the training set. We modify $M$ by adding a constant to the diagonal:
\[
M' = M + C \cdot I,
\]
where $C \in [0, \inf)$ is a regularization constant and $I$ is the identity matrix. Since the extreme values are standardized, the diagonal of $M$ is 1 and we can expect $C$ to have a consistent effect on each instance. The Mahalanobis distance $D_M$ is computed using
\[
D_M(x) = \sqrt{(x - \mu)^\mathsf{T} (M + C \cdot I)^{-1} (x - \mu)}.
\]
The value $C=0$, corresponds to the standard Mahalanobis distance, that represents  the likelihood of a multivariate normal distribution. If $C$ is large, the regularized covariance matrix $M'$ and the corresponding inverse covariance matrix $(M')^{-1}$ are dominated by the constant on the diagonal and $D_M$ is approximately proportional to the $L^2$ distance. This corresponds to the likelihood of a normal distribution where each of the extreme values are independent. The regularizer $C$ allows us to control the definition of in-distribution data. Algorithm 1 presents the pseudo-code of the \texttt{XOOD-M} algorithm. As can be seen from the pseudo-code, \texttt{XOOD-M} has minimal computational overhead. The only features that are computed on the training data are mean and covariance on the extreme value feature set, which is low dimensional with only $2\times r$ features, where $r$ is the number of activation layers in the network.

\begin{algorithm}
\caption{\textbf{Algorithm 1:} XOOD-M}\label{alg:m}
\begin{algorithmic}
\Require Training image and label set $<X_t, y_t>$ , Pretrained classifier $f_\theta$, Regularization $C$
\Ensure Mean $\mu$ and Covariance $K$ of OOD features

\State \texttt{Predict} $\hat{y}_t = f_\theta(X_t)$
\State $X_{t}$ = $X_{t, \hat{y}_t==y_t}$
\For{\texttt{each input $X'_{t,j}$ to activation layer $j$ in} $f_\theta$ }
\State $X_{min,j}$, $X_{max,j}$ = \texttt{FeatureExtraction}($X'_{t,j}$)
\State $X_{j}^{P}$, $T_{j}^{P}$ = \texttt{fit.PowerTransform}($X_{min,j}$, $X_{max,j}$)
\EndFor

\State $\mu$ = \texttt{Mean}($X_{j}^{P}$)
\State $M$ = \texttt{Covariance}($X_{j}^{P}$)

\end{algorithmic}
\end{algorithm}

\subsubsection{Self-supervised Out-of-Distribution Detection: \texttt{XOOD-L}}\label{lr}

The extreme value feature set extracted from the training data can be used in a self-supervised setting for OOD detection. While the Mahalanobis distance-based method produces a set of distances from the in-distribution features, using a logisitic regression-based method it is possible to generate labels to classify points as out of OOD, as shown in Equation 1. Additionally, the calibrated probability scores of the logistic regression  model can directly be interpreted as the amount of confidence of the model based on the similarity of the test instance with instances seen during training. The logistic regression model predicts the likelihood of an instance being in-distribution based on the distribution of the extreme values in each activation layer. It uses the extreme value features obtained from the training data as well as \emph{distorted} calibration data  (see Section \ref{distortion}) for training the model. The labels of the instances are generated based on whether the pre-trained neural network model can correctly predict each instance or not: all correctly classified instances are assigned label 1 in the logistic regression training, while all incorrectly classified instances have label 0. The intuition behind this is that, OOD data should be further away from the in-distribution data than images where the model makes a meaningful prediction. Therefore, this logistic regression should generalize to OOD images and assign them predictions close to 0. As shown in Figure \ref{max_example}, sometimes the distorted data is mapped to both sides of the in-distribution mode of the extreme value distribution. To enable the logistic regression to handle this, we split up the extreme values for each layer around the mean of the in-distribution data:
\[
\begin{cases}
m_i^+ = \text{relu}(m_i - \bar{m}_i) \\
m_i^- = \text{relu}(-m_i + \bar{m}_i) \\
\end{cases}
\]
where $m_i$ is maximum or minimum for activation layer $i$ and $\bar{m}_i$ is the mean of on the in-distribution data. Lastly, before applying the logistic regression, each of the split up extreme values are scaled to zero mean and unit variance to ensure consistent regularization. To avoid overfitting to the logistic regression training data, we apply $\ell_2$-regularization which is tuned through k-fold cross validation where each fold excludes one of the distortion types. Algorithm 2 describes the pseudo-code of the \texttt{XOOD-L} algorithm. 

\begin{algorithm}
\caption{\textbf{Algorithm 2:} XOOD-L}\label{alg:lr}
\begin{algorithmic}
\Require Training image and label set $<X_t, y_t>$, Calibration image and label set $<X_c, y_c>$, 
\Require Number of distortions to be applied $k$, Distortion parameters $d$, Pretrained classifier $f_\theta$
\Ensure Logistic regression model $g(w,\lambda)$ for training module

\State \texttt{Predict} $\hat{y}_t = f_\theta(X_t)$
\State $X_{t}$ = $X_{t, \hat{y}_t==y_t}$
\For{\texttt{each input $X'_{t,j}$ to activation layer $j$ in} $f_\theta$ }
\State $X_{min,j}$, $X_{max,j}$ = \texttt{FeatureExtraction}($X'_{t,j}$)
\State $X_{j}^{P}$, $T_{j}^{P}$ = \texttt{fit.PowerTransform}($X_{min,j}$, $X_{max,j}$)
\EndFor
\For{$i \in$ {$1, \dots, k$}}
        \State $X_{D_i}$ = \texttt{Distortion}($X_c, d_i$)
\EndFor
\State $Z$ = $[X_C, X_{D_1}, X_{D_2}, \dots, X_{D_k}]$
\For{\texttt{each input $Z'_{j}$ to activation layer $j$ in} $f_\theta$}
\State $Z_{min,j}$, $Z_{max,j}$ = \texttt{ExtremeValueExtraction}($Z'_{j}$)
\State $Z_j^{P}$ = $T_{j}^{P}$.\texttt{PowerTransform}($Z_{min,j}$, $Z_{max,j}$)
\EndFor
\State \texttt{Predict} $\hat{y}_c = f_\theta(Z)$
\State $LR_x$ = $[Z_{min,1}$ $Z_{max,1} Z_{min,2}$ $Z_{max,2} Z_{min,3}$ $Z_{max,3} \dots ]$
\If{$\hat{y}_c==y_c$}
\State $LR_{y}$ = 1
\Else
\State $LR_{y}$ = 0
\EndIf
\State $\lambda$ = \texttt{crossValidate.LogisticRegression}($LR_x, LR_y$) using each distorted data subset $X_{D_i}$ and $X_C$ as validation set to a $(k+1)$-fold cross validation 
\State $w$ = \texttt{fit.LogisticRegression}($LR_x, LR_y$, $\lambda$) 
\end{algorithmic}
\end{algorithm}

\subsubsection{Design Choices in Models} \label{design}

In both \texttt{XOOD-M} and \texttt{XOOD-L}, we make various choices in terms of the feature set, the distortions, and the regularization. In this section we discuss the rationale behind these choices.

\paragraph{}\textbf{Distortions}\label{distortion}
The distortions were designed to create a wide range of colors and textures, challenge the classifier in many different ways and produce wide and smooth distributions of extreme values. For satisfying these requirements, we used the following distortions in \texttt{XOOD-L}:

\begin{enumerate}
    \item\textbf{Geometric}: A combination of common geometric image augmentations: up to 90 degree rotation, up to 0.2 width and height shift, 50\% chance of horizontal flip, brightness range  between 0.2 and 2, zoom range from 0.9 to 1.1.
    \item\textbf{Mixup}: A convex combination of two images in the data set. The label is taken from the image with the highest weight. This produces images with less distinct features which lie near a decision boundary.
    \item\textbf{Gaussian Noise}: Additive Gaussian noise with mean 0 and variance ranging from 0 to 2. After adding noise we apply an affine transformation to each pixel value to create a wide range of brightness and contrast $x = ax + b$, where $a\in[1/8, 8]$ and $b \in [\min(0, 1-a), \max(0, 1-a)]$. $a$ and $b$ are constant for every image. 
    \item\textbf{Gaussian Blur}: Gaussian blur with mean 0 and variance ranging from 0.2 to 5. After blurring the images we apply the same affine transformation as for the Gaussian noise.
\end{enumerate}

All resulting images are clipped to have pixel values between 0 and 1. 

\begin{figure}[ht]
\centering
\includegraphics[width=1.\textwidth]{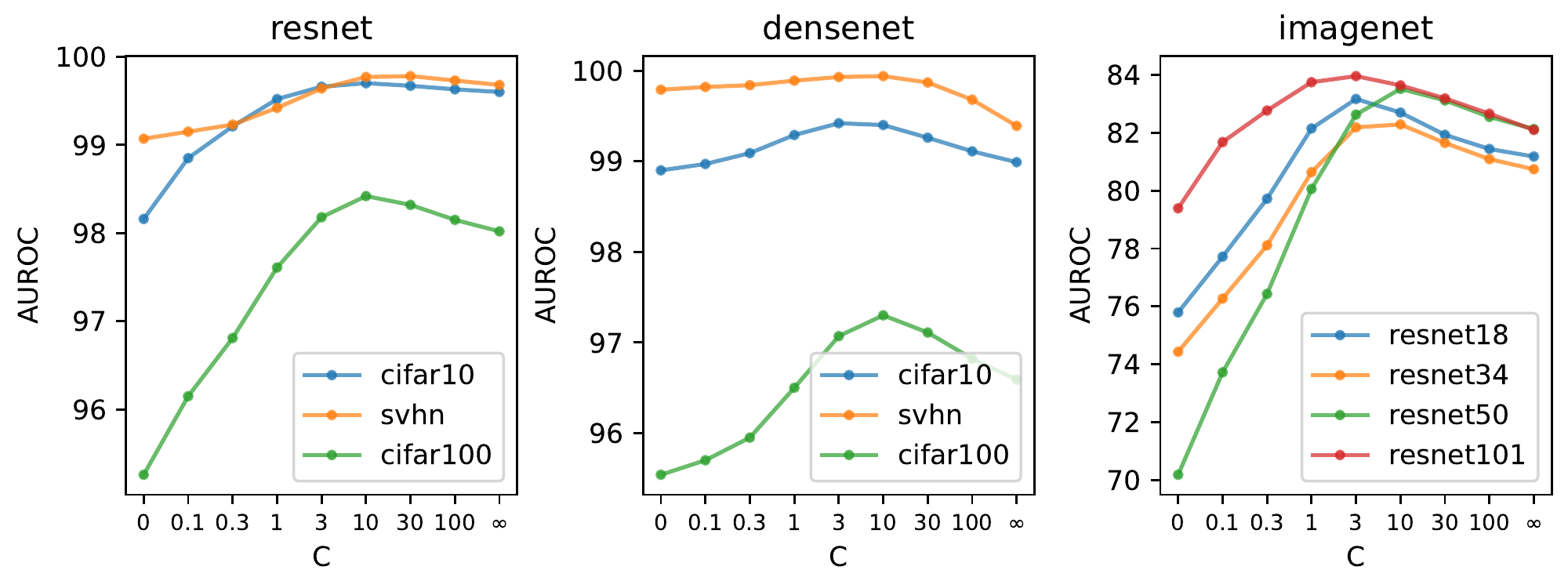}
\caption{AUROC-score of \texttt{XOOD-M} for various regularization values. The score was computed on the test set and the union of TinyImageNet (Crop), TinyImageNet (Resize), LSUN (Crop), LSUN (Resize), iSUN for CIFAR-10, CIFAR-100 and SVHN. For CIFAR-10 and CIFAR-100,  SVHN is included as OOD as well. For IMAGENET-1000, we used Places, SUN, iNaturalist and DTD, as curated by \cite{sun2021react}.}\label{fig:regularization}
\label{Mahala Tuning}
\end{figure}

\paragraph{}\textbf{Regularization}
While the $\ell_2$ regularization in the \texttt{XOOD-L} method is tuned through cross validation, the regularization parameter $C$ in \texttt{XOOD-M} cannot be learnt. So we experimented with a wide range of values for $C$. We found that a regularization $C=10$ worked well across all architectures and data sets. Figure \ref{fig:regularization} shows a comparison of detection performance with $C$ ranging from 0 to $\infty$, where $\infty$ corresponds to $L_2$-distance instead of Mahalanobis distance.

\paragraph{}\textbf{Feature Set} Although we based our algorithm on the extreme value distribution of the data in the activation layers, we experimented with a number of different features: only using the minimum or maximum, the sum; the percentage of positive values; $L_p$-norms; $L_p$-norms of $ReLU(x)$ and $ReLU(-x)$.  As one can see in Appendix B, neither replacing the extreme values by these features, nor adding these features while still using the extreme values improves the detection performance.

\section{Experiments}

In this section, we report the results obtained from running \texttt{XOOD-M} and \texttt{XOOD-L} on a variety of data sets on a number of performance metrics. The code is available at \cite{code}. We track performance with respect to the quality of the results as well as the efficiency of the methods. For quality of results, we measure performance based on the following metrics:

\begin{itemize}
    \item \textbf{AUROC:} The area under plot of true positive rate (TPR) versus false positive rate (FPR). A random detector has AUROC = 50\% and an ideal detector has AUROC = 100\%.
    \item \textbf{TNR (95\% TPR):} The probability that an OOD instance is correctly detected with a threshold that achieves a TPR of 95\%.
    \item \textbf{Detection Accuracy:} The maximum detection accuracy over all thresholds, assuming equal amounts of in-distribution and OOD instances. That is, $\max_T \{ \frac{p\left(f(x)>T| x \in \text{ID}\right) + p\left(f(x)\leq T| x \in \text{OOD}\right)}{2} \}$, where $f$ is the confidence score.
\end{itemize}
\def\XOOD{\texttt{XOOD }}
\subsection{Results}
We have tested the \texttt{XOOD} algorithms on ResNet and DenseNet for Cifar-10, Cifar-100, and SVHN using the same pretrained models as several other papers on OOD-detection. In Tables \ref{Table:ResNet} and \ref{Table:DenseNet}, we compare our algorithms with the baseline \cite{baseline}, ODIN \cite{odin}, Mahalanobis \cite{lee2018simple} and Gram \cite{sastry2020detecting}. Even though ODIN and Mahalanobis are trained on a sample of the OOD-data, \texttt{XOOD} outperforms them in most cases. And even though Gram is far more computationally complex, \XOOD outperforms it in most experiments. The one case where \XOOD struggles is CIFAR-100 vs CIFAR-10. However, this is not necessarily bad, because the datasets are very similar. For example, CIFAR-10 includes cars, trucks and dogs, while CIFAR-100 contains pickup trucks, busses and wolfs. If   all of these  are detected as OOD, it limits the generalization ability of the model.
\input{gram_table}

To summarize the performance of the various methods, we computed the mean scores across all metrics for each OOD detection task, except the CIFAR-10$-$CIFAR-100 in-distribution-OOD combination, given their overlap. We observe that \texttt{XOOD-L} and \texttt{XOOD-M} have the best FPR (95\% TPR) across all architectures, beating the state-of-the-art Gram method by approximately 50\%. For details, please refer to the Appendix D (Table \ref{fpr-table}).

% FPR (95\% TPR) over all datasets except CIFAR-10 vs CIFAR-100 and found that the three best results for ResNet34 were Gram (7.1\%), \texttt{XOOD-L} (5.2\%) and \texttt{XOOD-M} (3.5\%), and for DenseNet they were Gram (6.9\%), \texttt{XOOD-L} (3.6\%) and \texttt{XOOD-M} (5.1\%). For ResNet34, \texttt{XOOD-M} reduce the FPR by 51\% and for DenseNet \texttt{XOOD-L} reduce the FPR by 48\%.
\subsection{Inference Time} \label{inference}
In real-world applications of OOD-detection, the inference time is often crucial, yet this has largely been neglected in research. We measured the inference time of a ResNet and DenseNet on cifar10 and cifar100 for the baseline \cite{baseline}, Mahalanobis \cite{lee2018simple}, Gram \cite{sastry2020detecting} and \XOOD. We chose to compare with Mahalanobis and Gram since they have good detection performance and use similar feature extraction schemes as \XOOD. The results are shown in Tables \ref{Table: Time ResNet} and \ref{Tabel:Time DenseNet}. While \XOOD has an overhead of about 30\%, Mahalanobis and Gram impose more than a 10-fold increase in inference time. Mohseni et al. \cite{shifting_tranformation} compare the inference time of Outlier Exposure \cite{Hendrycks19}, Geometric OOD \cite{golan2018deep}, shifting tranformation learning \cite{shifting_tranformation}, SSD \cite{sehwag2021ssd}, CSI \cite{tack2020csi}, CSI-ens \cite{tack2020csi}, and Gram \cite{sastry2020detecting}. All  these algorithms have overhead above 100\% except Outlier Exposure, which doesn't have any overhead at all, but it is not really comparable to \XOOD since it requires fitting on OOD data.

The tables \ref{Table:ResNet} and \ref{Table:DenseNet} display the time it takes to compute the classification and OOD detection of 10000 images consisting of Gaussian noise on a single \textit{NVIDIA Tesla V100} GPU and one \textit{Intel Xeon Silver 4216} CPU with a batch size of 128, averaged over 10 runs.  We define overhead as $\frac{T-T_B}{T_B}$, where $T$ is the average inference time and $T_B$ is the average inference time of the baseline. Code for Mahalanobis \cite{lee2018simple} and Gram \cite{sastry2020detecting} was modified to enable this comparison. The computations in these algorithms do not depend on the content of the images and Gaussian noise was simply used for convenience.

\begin{table}[!htb]
\caption{\label{Table: Time ResNet}
Average inference time measured in seconds of various OOD-detection algorithms on ResNet34. ± indicates the 99\%-confidence interval of the mean.}

\begin{subtable}[h]{0.48\textwidth}
\caption{CIFAR-10}
\begin{tabular}{lrr}
\toprule
{} & Inference Time & Overhead \\
\midrule
Baseline    &            1.45 ± 0.0041 &       0\% \\
XOOD-L     &            1.79 ± 0.0575 &      24\% \\
XOOD-M     &            1.99 ± 0.0163 &      37\% \\
Mahalanobis &           20.25 ± 0.0413 &    1298\% \\
Gram        &           28.70 ± 0.6036 &    1880\% \\
\bottomrule
\end{tabular}
\label{table3}
\end{subtable}
\begin{subtable}[h]{0.48\textwidth}
\caption{CIFAR-100}
\begin{tabular}{lrr}
\toprule
{} & Inference Time & Overhead \\
\midrule
Baseline    &            1.45 ± 0.0069 &       0\% \\
XOOD-L     &            1.83 ± 0.0913 &      26\% \\
XOOD-M     &             2.0 ± 0.0177 &      38\% \\
Mahalanobis &           28.02 ± 0.4577 &    1834\% \\
Gram        &           28.49 ± 0.1467 &    1866\% \\
\bottomrule
\end{tabular}
\label{table4}
\end{subtable}
\end{table}

\begin{table}[!htb]
\caption{\label{Tabel:Time DenseNet}
Average inference time measured in seconds of various OOD-detection algorithms on DenseNet. ± indicates the 99\%-confidence interval of the mean.}
\begin{subtable}[h]{0.48\textwidth}
\caption{CIFAR-10}
\begin{tabular}{lrr}
\toprule
{} & Inference Time & Overhead \\
\midrule
Baseline    &            2.24 ± 0.0102 &       0\% \\
XOOD-L     &            2.90 ± 0.0497 &      29\% \\
XOOD-M     &            3.12 ± 0.0480 &      39\% \\
Mahalanobis &           30.11 ± 0.2802 &    1246\% \\
Gram        &           63.01 ± 0.7629 &    2717\% \\
\bottomrule
\end{tabular}
\end{subtable}
\begin{subtable}[h]{0.48\textwidth}
\caption{CIFAR-100}
\begin{tabular}{lrr}
\toprule
{} & Inference Time & Overhead \\
\midrule
Baseline    &            2.23 ± 0.0056 &       0\% \\
XOOD-L     &            2.91 ± 0.0849 &      31\% \\
XOOD-M     &            3.08 ± 0.0188 &      39\% \\
Mahalanobis &           36.88 ± 0.6617 &    1557\% \\
Gram        &           61.46 ± 0.6788 &    2662\% \\
\bottomrule
\end{tabular}
\end{subtable}
\end{table}

\subsection{Further Experiments}
Additionally, we compare the performance of \XOOD with Energy \cite{liu2020energy} and ReAct \cite{sun2021react} on DenseNet and ResNet34 for CIFAR-10, CIFAR-100 and SVHN, as well as ResNet18, ResNet34, ResNet50 and ResNet101 for Imagenet1000. A table with these results can be found in Appendix A. \XOOD  outperforms the other algorithms on CIFAR-10, CIFAR-100 and SVHN. ReAct performs the best on Imagenet1000, but does not significantly improve over the baseline on the other data sets. It also struggles with the Gaussian noise data set on Imagenet1000 and gets 0\% TPR at 95\% TNR for ResNet34 and ResNet50. \texttt{XOOD-L}  outperforms the baseline and Energy on almost all combinations in in-distribution and OOD data sets. 

\section{Related Work}
The last few years have seen a tremendous amount of work in OOD detection \cite{yang2021generalized,odin, sastry2020detecting,sun2021react,ren2019likelihood,odin,lee2018simple,ren2019likelihood,devries2018learning,liu2020energy,vyas2018out,yu2019unsupervised,lin2021mood,mohseni2020self,huang2021mos,kumar2021calibrated,hendrycks2019scaling}. However, as argued in \cite{tajwar2021no}, no single algorithm can be considered the state-of-the-art. In \cite{tajwar2021no}, the authors tested the algorithms in \cite{odin,baseline,lee2018simple} on three benchmark in-distribution datasets (CIFAR-10, CIFAR-100, SVHN) and seven benchmark OOD datasets under standardized conditions. They found inconsistent performance of these algorithms across all the datasets; in fact, there was no algorithm that was consistently outperforming others across all the datasets. As shown in Tables \ref{Table:DenseNet}, \ref{Table:ResNet}, for DenseNet and ResNet34 architectures,  both   \texttt{XOOD-L} and \texttt{XOOD-M}   outperform \cite{odin,baseline,lee2018simple} for CIFAR-10 (with LSUN-C, LSUN-R, SVHN, TinyImageNet-C, TinyImageNet-R, and iSun as OOD). Similar is the case with most other in-distribution-OOD combinations. Unlike algorithms like \cite{odin,lee2018simple,liu2020energy}, neither \texttt{XOOD-M} nor \texttt{XOOD-L} require access to the OOD data during training time. In addition, for the architectures considered, both XOOD-L and XOOD-M outperform \cite{sastry2020detecting} for almost all in-distribution-OOD combinations. Mohseni et al. \cite{shifting_tranformation} compare different OOD detection algorithms in terms of inference time. All the algorithms considered, that are not allowed to train on OOD data,  have overhead above 100\%. Both the algorithms in \cite{sastry2020detecting,lee2018simple} involve matrix multiplication which has at least quadratic complexity. Since, the number of features within a neural network is large, matrix multiplication operations as used in \cite{sastry2020detecting,lee2018simple} are prohibitively expensive. In case of \texttt{XOOD-M}, Mahalanobis distance is computed on the extreme values which has dimension twice the depth of the model (typically at most 1000); hence Mahalanobis distance can be computed efficiently. This explains why  both our algorithms \texttt{XOOD-L} and \texttt{XOOD-M} outperform \cite{sastry2020detecting} and \cite{lee2018simple} in terms of inference time by an order of magnitude. 
In Table \ref{reacttable} in Appendix A, we show that ReAct \cite{sun2021react} is having difficulty distinguishing Gaussian noise from in-distribution ImageNet (Resnet 34 and ResNet 50). Table \ref{reacttable} also shows that both \texttt{XOOD-L} and \texttt{XOOD-M} are robust to Gaussian and uniform noise. Additionally, Table \ref{reacttable} shows that the \texttt{XOOD} framework (in particular \texttt{XOOD-L}) outperforms \cite{liu2020energy} on all but one (in-distribution CIFAR-10, OOD CIFAR-100) in-distribution-OOD data set combinations for all the architectures considered.   ReAct performs the best on Imagenet1000, but does not significantly improve over the baseline on the other data sets.
% \section{Discussion}
% \subsection{Why does it work?}
% Both algorithms \texttt{XOOD-L} and \texttt{XOOD-M}  rely on computation of global extrema of input features before each activation layer.
% Both algorithms are based on the observation that  the per layer distribution of extrema is consistently unimodal on in-distribution data. They look for deviations from this behavior. 
% Just listing some thought:

% - Looking for deviations from the behaviour seen on in-distribution data.
% - Batch normalization makes in-distribution data statistically well behaved. The extreme values detect deviations from this behaviour.
% - The relu non-linearity at 0 is essential for the expressive power of the networks. The extreme values indicate how far away from this non-linearity the activations get. The asymmetry of the relu makes it meaningful to use maximum and minimum rather than the maximum absolute value.

% \subsection{Which algorithm should you choose?}
% Mahalanobis:
% + Unsupervised.
% + Doesn't rely on anything image specific.
% + Quick to set up.
% + Better on some datasets: more robust since it only assigns high confidence to a compact set.
% - Regularization parameter might need manual tuning.
% Logistic Regression:
% + Customizable.
% + Tunes it's own regularization.
% + Better on some datasets: it can learn whether to use put decision boundries on both directions from the in-distribution data, or only one direction. 
% - Requires distorted images.

\section{Conclusions}
We presented the \XOOD framework (Extreme Value-based OOD Detection) that comprises two new, efficient and accurate algorithms for OOD detection for the image classification:  the unsupervised \texttt{XOOD-M}, and the self-supervised \texttt{XOOD-L}. They rely on the signals captured by the extreme values of the data in the activation layers of the neural network to distinguish between in distribution and out of distribution instances. On most benchmark in-distribution-OOD dataset combinations, both \texttt{XOOD-L} and \texttt{XOOD-M} outperform state-of-the-art algorithms for standard architectures.
\section{Limitations of the Work}\label{limit}
The current work is focused on image classification. In future, we would like to extend the work to language models as well as graph neural networks, as well as test the \XOOD algorithms in a more general confidence and calibration context.
\section{Impact}\label{impact}
When encountered with OOD examples at inference time, neural networks can make erroneous decisions rather than issuing a warning that  their decisions in these cases  cannot be trusted. Such behavior prevents their deployment in mission-critical application. The \XOOD framework can help improve the trustworthiness of deep neural networks through efficient and accurate OOD detection. We do not anticipate any negative social consequences of our work. 
% \section*{Acknowledgments}
% %\begin{ack}
% Use unnumbered first level headings for the acknowledgments. All acknowledgments
% More information about this disclosure can be found at: \url{https://neurips.cc/Conferences/2022/PaperInformation/FundingDisclosure}.

% Do {\bf not} include this section in the anonymized submission, only in the final paper. You can use the \texttt{ack} environment provided in the style file to autmoatically hide this section in the anonymized submission.
% \end{ack}

\bibliographystyle{unsrt}
\bibliography{references}
\newpage

\section*{A: Additional Experiments}\label{appendixA}

\input{new_react_table}

\section*{B: Feature Selection}\label{appendixB}
In this section, we report experimentation with a number of different features: only using the minimum or maximum, the percentage of positive values (positivity); the sum; $L_p$-norms; $L_p$-norms of $\text{ReLU}(x)$ and $\text{ReLU}(-x)$ (Split Lp).  As one can see in Tables \ref{Table: Feature Selection} and \ref{Table: Feature Addition}, neither replacing the extreme values by these features, nor adding these features while still using the extreme values improves the detection performance.
\input{feature_selection}

\input{feature_addition}

\newpage
\section*{C: Extreme Value Distributions} \label{appendixC}

\begin{figure}[h]
    \centering
    \includegraphics[width=.6\textwidth]{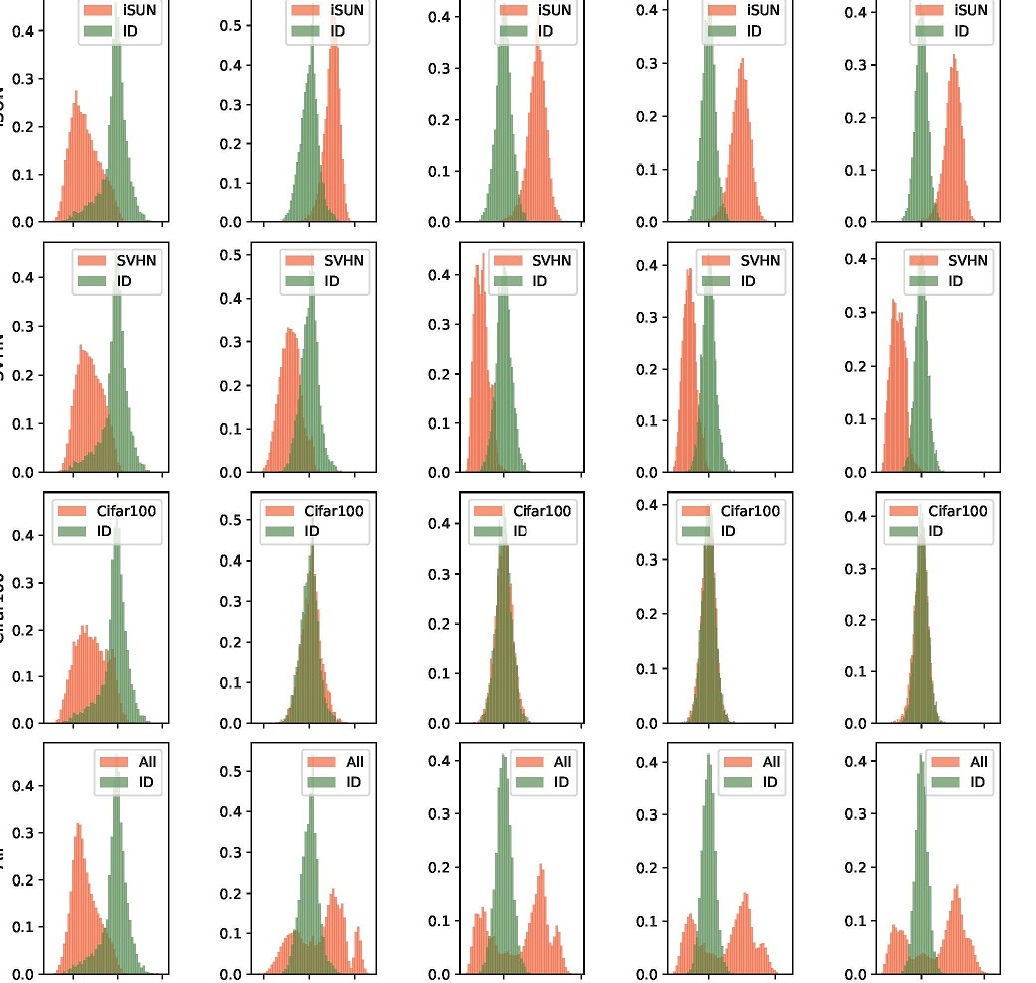}
    \caption{Distribution of maximum values before an activation layer inside ResNet34. Green represents CIFAR-10 (In-distribution (ID)) and  Orange represents OOD. Note how in-distribution data lies in a single mode, while out-of-distribution data can be on either side.}
    \label{CIFARmax}
\end{figure}
\begin{figure}[h]
    \centering
    \includegraphics[width=.6\textwidth]{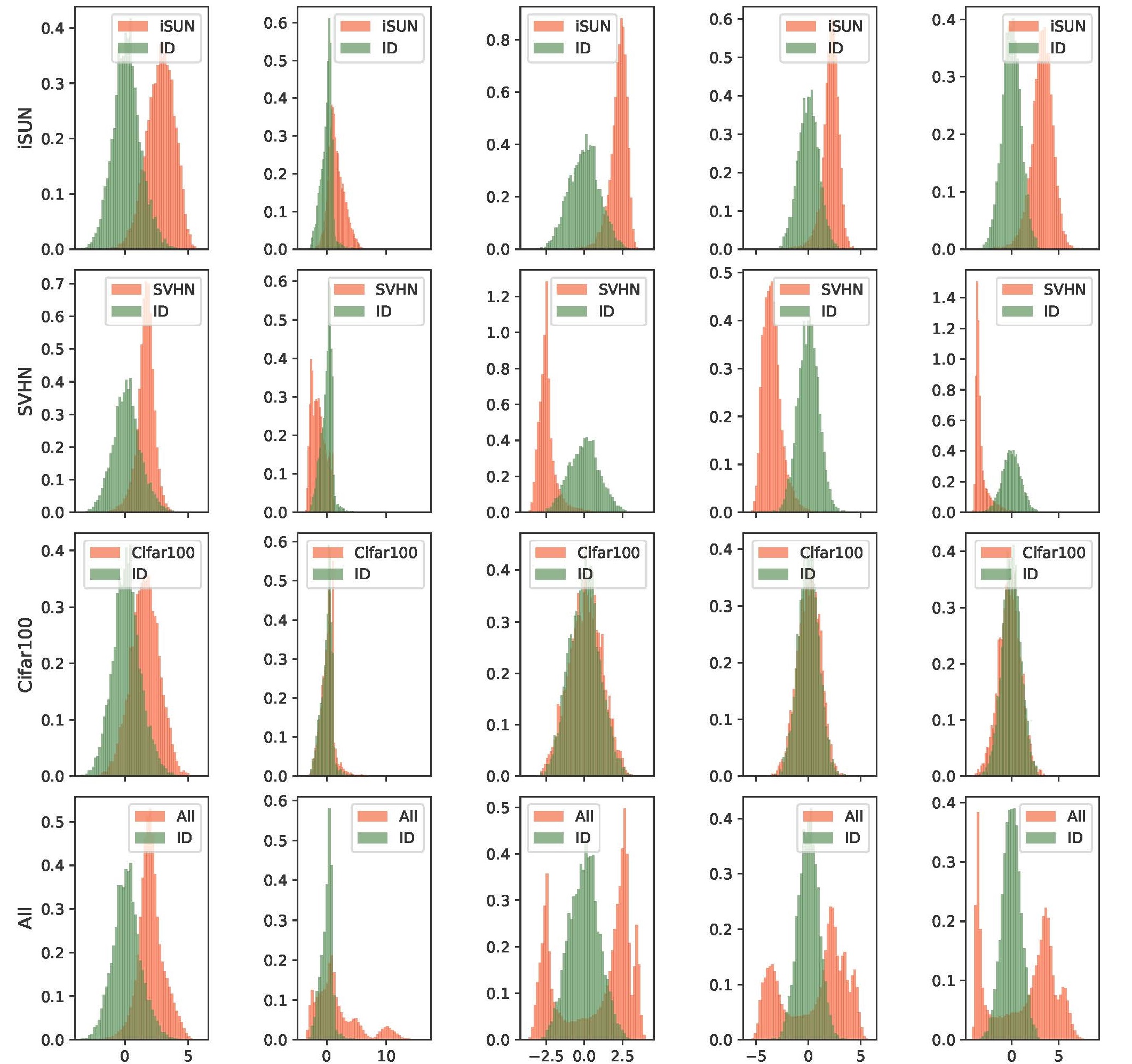}
    \caption{Distribution of minimum values before an activation layer inside ResNet34. Green represents CIFAR-10 (In-distribution (ID)) and  Orange represents OOD. Note how in-distribution data lies in a single mode, while out-of-distribution data can be on either side.}
    \label{CIFARmin}
\end{figure}
\newpage
\section*{D: False Positive Rate}

To summarize the performance of the various methods in Table \ref{Table:DenseNet} and \ref{Table:ResNet}, we computed the mean FPR (95\% TPR) for each OOD detection task, except the CIFAR-10$-$CIFAR-100 and CIFAR-100$-$CIFAR-10 in-distribution-OOD combinations, given their overlap. These averages are shown in Table \ref{fpr-table}. Note that \texttt{XOOD-L} and \texttt{XOOD-M} have the best average scores for both DenseNet and ResNet. For DenseNet, \texttt{XOOD-L} reduced the FPR by 48\% compared to the best other method (Gram), and for ResNet34  \texttt{XOOD-M} reduced the FPR by 51\% compared to the best other method (Gram).

\input{fpr_table}

\end{document}

%% file: gram_table.tex
\begin{table}[ht]
\caption{\textbf{DenseNet} detection scores for Baseline/Odin/Mahalanobis/Gram/XOOD-L/XOOD-M. Results for other algorithms are from  Sastry and Oore \cite{gram}.} \label{Table:DenseNet}
\resizebox{\textwidth}{!}{
\begin{tabular}{llccc}
\toprule
     &      &                                    TNR (95\% TPR) &                                                     AUROC &                                   Detection Acc. \\
In Dist & Out Dist &                                                  &                                                           &                                                  \\
\midrule
CIFAR-10 & CIFAR-100 &           40.3/\textbf{53.1}/14.5/26.7/47.6/19.9 &                    89.3/\textbf{90.2}/58.5/72.0/87.0/70.7 &           \textbf{82.9}/82.7/57.2/67.3/78.6/64.8 \\
     & LSUN-C &           51.8/70.6/48.2/88.4/\textbf{96.9}/91.7 &                    92.9/93.6/80.2/97.5/\textbf{99.3}/98.4 &           86.9/86.4/75.6/92.0/\textbf{96.1}/93.6 \\
     & LSUN-R &           66.6/96.2/97.2/99.5/99.4/\textbf{99.7} &  95.4/99.2/99.3/\textbf{99.9}/\textbf{99.9}/\textbf{99.9} &  90.3/95.7/96.3/\textbf{98.6}/98.5/\textbf{98.6} \\
     & SVHN &           40.2/86.2/90.8/96.1/96.5/\textbf{97.2} &                    89.9/95.5/98.1/99.1/99.2/\textbf{99.3} &           83.2/91.4/93.9/95.9/96.2/\textbf{96.7} \\
     & TinyImgNet-C &           56.7/87.0/84.2/96.7/\textbf{98.6}/97.8 &                    93.8/97.6/95.3/99.3/\textbf{99.7}/99.5 &           88.1/92.3/89.9/96.1/\textbf{97.7}/96.6 \\
     & TinyImgNet-R &           58.9/92.4/95.0/98.8/99.1/\textbf{99.2} &           94.1/98.5/98.8/99.7/\textbf{99.8}/\textbf{99.8} &           88.5/93.9/95.0/97.9/\textbf{98.2}/97.8 \\
     & iSUN &           62.5/93.2/95.3/99.0/\textbf{99.5}/99.4 &                    94.7/98.7/98.9/99.8/\textbf{99.9}/99.8 &           89.2/94.3/95.2/97.9/\textbf{98.5}/98.1 \\
CIFAR-100 & CIFAR-10 &           \textbf{18.9}/16.8/ 7.7/10.6/10.0/ 1.7 &                    \textbf{75.9}/74.2/60.1/64.2/65.0/43.1 &           \textbf{69.7}/68.6/57.8/60.4/61.8/50.0 \\
     & LSUN-C &           28.6/57.8/42.1/65.5/\textbf{84.9}/68.4 &                    80.2/91.4/81.7/91.4/\textbf{97.4}/92.6 &           72.7/83.3/74.0/83.6/\textbf{91.3}/85.0 \\
     & LSUN-R &           17.6/41.2/91.4/97.2/94.3/\textbf{97.9} &           70.8/85.5/98.0/\textbf{99.3}/99.0/\textbf{99.3} &           64.9/77.1/93.9/96.4/94.8/\textbf{96.7} \\
     & SVHN &           26.7/70.6/82.5/\textbf{89.3}/88.8/87.9 &                    82.7/93.8/97.2/97.3/\textbf{97.7}/97.1 &           75.6/86.6/91.5/\textbf{92.4}/92.3/91.9 \\
     & TinyImgNet-C &           24.6/51.0/60.1/89.0/\textbf{95.4}/89.2 &                    76.2/88.3/88.8/97.7/\textbf{99.1}/97.5 &           69.0/80.2/81.6/92.5/\textbf{95.2}/92.7 \\
     & TinyImgNet-R &           17.6/42.6/86.6/95.7/95.7/\textbf{96.1} &                    71.7/85.2/97.4/99.0/\textbf{99.2}/98.9 &           65.7/77.0/92.2/95.5/95.4/\textbf{95.6} \\
     & iSUN &           14.9/37.4/87.0/95.9/94.0/\textbf{96.3} &           69.5/84.5/97.4/\textbf{99.0}/98.9/\textbf{99.0} &           63.8/76.4/92.4/95.6/94.6/\textbf{95.7} \\
SVHN & CIFAR-10 &           69.3/71.7/96.8/80.4/\textbf{99.7}/98.4 &                    91.9/91.4/98.9/95.5/\textbf{99.8}/99.5 &           86.6/85.8/95.9/89.1/\textbf{98.3}/97.0 \\
     & LSUN-R &  77.1/81.1/99.9/99.5/\textbf{100.}/\textbf{100.} &           94.1/94.5/99.9/99.8/\textbf{100.}/\textbf{100.} &  89.1/89.2/99.3/98.6/\textbf{99.8}/\textbf{99.8} \\
     & TinyImgNet-R &  79.8/84.1/99.9/99.1/\textbf{100.}/\textbf{100.} &           94.8/95.1/99.9/99.7/\textbf{100.}/\textbf{100.} &           90.2/90.4/98.9/97.9/99.5/\textbf{99.6} \\
     & iSUN &  78.3/82.2/99.9/99.4/\textbf{100.}/\textbf{100.} &           94.4/94.7/99.9/99.8/\textbf{100.}/\textbf{100.} &           89.6/89.7/99.2/98.3/99.7/\textbf{99.8} \\
\bottomrule
\end{tabular}
}
\end{table}

\begin{table}[ht]
\caption{\textbf{ResNet34} detection scores for Baseline/Odin/Mahalanobis/Gram/XOOD-L/XOOD-M. Results for other algorithms are from  Sastry and Oore \cite{gram}.} \label{Table:ResNet}
\resizebox{\textwidth}{!}{%
\begin{tabular}{llccc}
\toprule
     &      &                                    TNR (95\% TPR) &                                                     AUROC &                          Detection Acc. \\
In Dist & Out Dist &                                                  &                                                           &                                         \\
\midrule
CIFAR-10 & CIFAR-100 &           33.3/42.0/41.6/32.9/\textbf{43.6}/32.0 &                    86.4/85.8/88.2/79.0/\textbf{88.8}/80.1 &  80.4/78.6/81.2/71.7/\textbf{81.7}/73.7 \\
     & LSUN-C &           48.6/62.0/81.3/89.8/\textbf{97.0}/94.6 &                    91.9/91.2/96.7/97.8/\textbf{99.4}/99.0 &  86.3/82.4/90.5/92.6/\textbf{96.0}/94.8 \\
     & LSUN-R &           49.8/82.1/98.8/99.6/99.1/\textbf{99.9} &                    91.0/94.1/99.7/99.9/99.8/\textbf{100.} &  85.3/86.7/97.7/98.6/97.8/\textbf{99.2} \\
     & SVHN &           50.5/70.3/87.8/97.6/96.4/\textbf{98.4} &                    89.9/96.7/99.1/99.5/99.3/\textbf{99.7} &  85.1/91.1/95.8/96.7/96.1/\textbf{97.6} \\
     & TinyImgNet-C &           46.4/68.7/92.0/96.7/98.0/\textbf{99.1} &                    91.4/93.1/98.6/99.2/99.5/\textbf{99.8} &  85.4/85.2/93.9/96.1/96.7/\textbf{98.0} \\
     & TinyImgNet-R &           41.0/67.9/97.1/98.7/97.9/\textbf{99.5} &                    91.0/94.0/99.5/99.7/99.6/\textbf{99.9} &  85.1/86.5/96.3/97.8/96.9/\textbf{98.8} \\
     & iSUN &           44.6/73.2/97.8/99.3/99.1/\textbf{99.7} &                    91.0/94.0/99.5/99.8/99.8/\textbf{99.9} &  85.0/86.5/96.7/98.1/97.8/\textbf{98.8} \\
CIFAR-100 & CIFAR-10 &           19.1/18.7/\textbf{20.2}/12.2/11.7/ 9.3 &                    77.1/77.2/\textbf{77.5}/67.9/71.0/63.9 &  71.0/71.2/\textbf{72.1}/63.4/67.1/61.6 \\
     & LSUN-C &           18.7/44.1/64.8/64.8/\textbf{85.3}/76.2 &                    75.5/82.7/92.0/92.1/\textbf{97.5}/95.3 &  69.2/75.9/84.0/84.2/\textbf{91.4}/88.2 \\
     & LSUN-R &           18.8/23.2/90.9/96.6/90.2/\textbf{98.5} &                    75.8/85.6/98.2/99.2/98.3/\textbf{99.6} &  69.9/78.3/93.5/96.7/92.9/\textbf{97.4} \\
     & SVHN &           20.3/62.7/91.9/80.8/87.2/\textbf{92.1} &                    79.5/93.9/\textbf{98.4}/96.0/97.6/98.2 &  73.2/88.0/\textbf{93.7}/89.6/91.5/93.6 \\
     & TinyImgNet-C &           24.3/44.3/80.9/88.5/90.8/\textbf{95.0} &                    79.7/85.4/96.3/97.7/98.4/\textbf{98.9} &  72.5/78.3/89.9/92.2/93.2/\textbf{95.0} \\
     & TinyImgNet-R &           20.4/36.1/90.9/94.8/89.4/\textbf{97.9} &                    77.2/87.6/98.2/98.9/98.1/\textbf{99.5} &  70.8/80.1/93.3/95.0/92.6/\textbf{96.7} \\
     & iSUN &           16.9/45.2/89.9/94.8/89.6/\textbf{97.0} &                    75.8/85.5/97.9/98.8/98.3/\textbf{99.3} &  70.1/78.5/93.1/95.6/92.8/\textbf{96.2} \\
SVHN & CIFAR-10 &           78.3/79.8/\textbf{98.4}/85.8/98.1/96.1 &                    92.9/92.1/99.3/97.3/\textbf{99.5}/98.9 &  90.0/89.4/\textbf{96.9}/92.0/96.8/95.7 \\
     & LSUN-R &  74.3/77.3/\textbf{99.9}/99.6/99.5/\textbf{99.9} &  91.6/89.4/\textbf{99.9}/99.8/\textbf{99.9}/\textbf{99.9} &  89.0/87.2/\textbf{99.5}/98.5/98.3/98.9 \\
     & TinyImgNet-R &           79.0/82.0/\textbf{99.9}/99.3/99.7/99.8 &           93.5/92.0/\textbf{99.9}/99.7/\textbf{99.9}/99.8 &  90.4/89.4/\textbf{99.1}/97.9/98.4/98.6 \\
     & iSUN &           77.1/79.1/99.7/99.4/99.8/\textbf{99.9} &           92.2/91.4/99.8/99.8/\textbf{99.9}/\textbf{99.9} &  89.7/89.2/98.3/98.1/98.7/\textbf{98.9} \\
\bottomrule
\end{tabular}
}
\end{table}

%% file: new_react_table.tex
\begin{table}[H]
\caption{Baseline/Energy/React/XOOD-L/XOOD-M.}
\resizebox{\textwidth}{!}{

%%%%%%%%%%%%%%%%%%%%%%%%%%%%%%%%%%%%%%%%%%%%%%%%%%%%%%%%%%%%

\begin{tabular}{llllll}
\toprule
         &          &      &                                                 TNR (95\% TPR) &                              Detection Acc. &                                       AUROC \\
\midrule
imagenet & resnet18 & Uniform &   2.5/\textbf{100.}/\textbf{100.}/\textbf{100.}/\textbf{100.} &           86.9/99.1/99.7/\textbf{100.}/99.9 &  86.9/99.7/99.9/\textbf{100.}/\textbf{100.} \\
         &          & Gaussian &                     0.0/80.4/72.1/\textbf{100.}/\textbf{100.} &           76.0/95.3/94.4/\textbf{100.}/99.9 &  65.1/96.2/95.8/\textbf{100.}/\textbf{100.} \\
         &          & Places &                             23.9/34.8/\textbf{57.2}/37.1/34.2 &           70.7/75.9/\textbf{82.3}/76.7/70.7 &           78.0/83.6/\textbf{89.7}/84.4/77.1 \\
         &          & SUN &                             26.4/40.0/\textbf{65.5}/46.9/48.6 &           71.6/78.4/\textbf{85.3}/79.9/77.9 &           78.9/86.1/\textbf{92.7}/88.0/85.2 \\
         &          & iNaturalist &                             41.6/43.2/\textbf{69.9}/60.9/38.9 &           78.9/82.5/\textbf{87.4}/84.8/74.4 &           87.0/89.8/\textbf{94.6}/92.8/82.0 \\
         &          & DTD &                             28.6/46.7/60.8/56.9/\textbf{61.8} &           71.3/79.1/\textbf{83.6}/81.4/81.8 &           78.4/86.3/\textbf{91.4}/88.9/89.5 \\
         & resnet34 & Uniform &           86.4/\textbf{100.}/99.6/\textbf{100.}/\textbf{100.} &  94.6/98.1/97.3/\textbf{100.}/\textbf{100.} &  97.2/98.7/98.4/\textbf{100.}/\textbf{100.} \\
         &          & Gaussian &                    48.6/ 0.0/ 0.0/\textbf{100.}/\textbf{100.} &  92.5/89.8/89.1/\textbf{100.}/\textbf{100.} &  94.3/85.9/85.2/\textbf{100.}/\textbf{100.} \\
         &          & Places &                             26.1/37.2/\textbf{64.1}/38.6/34.2 &           71.4/77.0/\textbf{84.2}/77.8/69.8 &           79.2/84.6/\textbf{91.4}/85.6/75.9 \\
         &          & SUN &                             27.6/42.6/\textbf{73.6}/48.4/48.1 &           72.3/79.2/\textbf{87.5}/81.0/76.9 &           79.8/86.6/\textbf{94.0}/88.6/83.8 \\
         &          & iNaturalist &                             40.5/46.0/\textbf{75.9}/65.1/40.2 &           78.3/82.2/\textbf{88.8}/86.1/74.1 &           86.6/89.7/\textbf{95.7}/93.7/81.5 \\
         &          & DTD &                             29.9/44.9/55.3/61.5/\textbf{70.5} &           71.7/79.3/83.5/83.7/\textbf{85.1} &           79.1/86.5/91.2/91.0/\textbf{92.4} \\
         & resnet50 & Uniform &   3.6/\textbf{100.}/\textbf{100.}/\textbf{100.}/\textbf{100.} &  89.7/98.3/99.6/\textbf{100.}/\textbf{100.} &  89.8/98.9/99.8/\textbf{100.}/\textbf{100.} \\
         &          & Gaussian &                     0.0/ 0.0/ 0.0/\textbf{100.}/\textbf{100.} &  69.7/81.6/87.6/\textbf{100.}/\textbf{100.} &  53.2/75.5/85.8/\textbf{100.}/\textbf{100.} \\
         &          & Places &                             28.0/34.8/\textbf{66.6}/43.5/32.0 &           72.8/76.9/\textbf{85.0}/79.6/69.9 &           80.6/84.2/\textbf{92.0}/87.2/75.9 \\
         &          & SUN &                             31.0/41.9/\textbf{76.0}/54.4/47.8 &           74.0/79.4/\textbf{87.9}/82.7/77.6 &           81.7/86.8/\textbf{94.4}/90.4/84.8 \\
         &          & iNaturalist &                             46.8/46.3/\textbf{80.5}/71.2/44.8 &           79.9/83.9/\textbf{90.0}/88.3/76.9 &           88.4/90.7/\textbf{96.4}/95.1/84.6 \\
         &          & DTD &                             33.5/47.9/54.3/64.8/\textbf{71.0} &           73.0/79.7/82.8/84.1/\textbf{85.4} &           80.4/86.8/90.5/91.3/\textbf{92.9} \\
         & resnet101 & Uniform &           27.1/96.7/\textbf{100.}/\textbf{100.}/\textbf{100.} &  87.3/96.5/99.3/\textbf{100.}/\textbf{100.} &  90.8/97.2/99.6/\textbf{100.}/\textbf{100.} \\
         &          & Gaussian &            0.0/ 0.0/\textbf{100.}/\textbf{100.}/\textbf{100.} &  79.0/88.0/98.3/\textbf{100.}/\textbf{100.} &  70.6/83.9/98.2/\textbf{100.}/\textbf{100.} \\
         &          & Places &                             28.8/39.4/\textbf{63.7}/42.7/26.7 &           72.9/78.2/\textbf{84.2}/79.7/71.2 &           80.6/85.5/\textbf{91.3}/87.2/77.0 \\
         &          & SUN &                             32.1/46.8/\textbf{73.4}/51.0/41.3 &           73.9/80.6/\textbf{87.2}/82.6/78.3 &           81.4/87.7/\textbf{94.0}/89.9/85.2 \\
         &          & iNaturalist &                             42.9/41.6/\textbf{79.1}/61.9/32.5 &           77.4/81.1/\textbf{89.5}/86.3/75.7 &           86.2/88.3/\textbf{96.1}/93.4/82.8 \\
         &          & DTD &                             37.4/52.6/59.6/67.5/\textbf{68.6} &           73.4/80.8/83.5/85.4/\textbf{87.2} &           81.5/87.9/91.4/92.4/\textbf{94.0} \\
cifar10 & resnet & Uniform &                    72.7/82.0/90.2/\textbf{100.}/\textbf{100.} &  93.6/94.2/94.6/\textbf{100.}/\textbf{100.} &  96.1/96.2/96.8/\textbf{100.}/\textbf{100.} \\
         &          & Gaussian &                    90.8/98.8/98.1/\textbf{100.}/\textbf{100.} &  95.7/97.0/96.6/\textbf{100.}/\textbf{100.} &  97.5/98.0/98.2/\textbf{100.}/\textbf{100.} \\
         &          & TinyImageNet (Crop) &                             46.9/62.1/65.1/98.0/\textbf{99.1} &           85.5/86.5/86.5/96.7/\textbf{98.0} &           91.6/93.1/92.6/99.5/\textbf{99.8} \\
         &          & TinyImageNet (Resize) &                             45.1/59.5/63.9/97.9/\textbf{99.5} &           85.1/85.8/86.3/96.9/\textbf{98.8} &           91.1/92.5/92.4/99.6/\textbf{99.9} \\
         &          & LSUN (Crop) &                             49.1/65.7/62.5/\textbf{97.0}/94.6 &           86.4/87.5/85.1/\textbf{96.0}/94.8 &           92.0/93.9/90.9/\textbf{99.4}/99.0 \\
         &          & LSUN (Resize) &                             46.3/62.9/69.4/99.1/\textbf{99.9} &           85.5/86.7/88.2/97.8/\textbf{99.2} &           91.3/92.9/94.4/99.8/\textbf{100.} \\
         &          & iSUN &                             45.6/62.0/68.0/99.1/\textbf{99.7} &           85.2/86.3/87.9/97.8/\textbf{98.8} &           91.2/92.8/94.2/99.8/\textbf{99.9} \\
         &          & SVHN &                             33.3/47.9/37.9/96.4/\textbf{98.4} &           85.4/85.5/77.5/96.1/\textbf{97.6} &           90.1/91.3/83.3/99.3/\textbf{99.7} \\
         &          & Cifar100 &                    34.2/\textbf{44.0}/\textbf{44.0}/43.6/32.0 &           80.5/80.7/78.7/\textbf{81.7}/73.7 &           86.6/87.3/84.4/\textbf{88.8}/80.1 \\
         & densenet & Uniform &           78.0/96.7/\textbf{100.}/\textbf{100.}/\textbf{100.} &  94.8/96.3/98.1/\textbf{100.}/\textbf{100.} &  96.7/97.4/99.2/\textbf{100.}/\textbf{100.} \\
         &          & Gaussian &  89.3/\textbf{100.}/\textbf{100.}/\textbf{100.}/\textbf{100.} &  95.4/97.9/98.4/\textbf{100.}/\textbf{100.} &  97.7/98.9/99.5/\textbf{100.}/\textbf{100.} \\
         &          & TinyImageNet (Crop) &                             57.4/84.1/68.6/\textbf{98.6}/97.8 &           88.2/91.6/87.6/\textbf{97.7}/96.6 &           94.0/97.1/94.1/\textbf{99.7}/99.5 \\
         &          & TinyImageNet (Resize) &                             60.1/86.2/68.7/99.1/\textbf{99.2} &           88.8/92.3/87.3/\textbf{98.2}/97.8 &  94.3/97.5/94.0/\textbf{99.8}/\textbf{99.8} \\
         &          & LSUN (Crop) &                             52.4/76.1/66.4/\textbf{96.9}/91.7 &           87.2/89.8/88.4/\textbf{96.1}/93.6 &           93.2/96.0/94.4/\textbf{99.3}/98.4 \\
         &          & LSUN (Resize) &                             67.3/92.3/83.1/99.4/\textbf{99.7} &           90.5/93.9/91.4/98.5/\textbf{98.6} &  95.6/98.4/96.9/\textbf{99.9}/\textbf{99.9} \\
         &          & iSUN &                             63.8/89.3/78.4/\textbf{99.5}/99.4 &           89.5/93.1/90.0/\textbf{98.5}/98.1 &           95.0/98.0/96.1/\textbf{99.9}/99.8 \\
         &          & SVHN &                             40.9/52.9/65.3/96.5/\textbf{97.2} &           83.4/83.5/88.1/96.2/\textbf{96.7} &           90.1/91.1/93.8/99.2/\textbf{99.3} \\
         &          & Cifar100 &                             41.7/\textbf{55.3}/49.3/47.6/19.9 &           83.2/\textbf{83.6}/82.2/78.6/64.8 &           89.6/\textbf{90.7}/89.2/87.0/70.7 \\
svhn & resnet & Uniform &                    85.9/86.7/71.9/\textbf{100.}/\textbf{100.} &           93.0/92.5/84.7/99.5/\textbf{99.7} &  96.1/95.6/89.4/\textbf{100.}/\textbf{100.} \\
         &          & Gaussian &                    86.1/87.3/75.2/\textbf{100.}/\textbf{100.} &           93.0/92.7/86.1/99.7/\textbf{99.9} &  96.3/96.0/90.5/\textbf{100.}/\textbf{100.} \\
         &          & TinyImageNet (Crop) &                    81.3/83.0/74.4/\textbf{99.7}/\textbf{99.7} &           91.1/90.7/85.5/\textbf{98.7}/98.3 &           94.2/93.8/88.7/\textbf{99.9}/99.8 \\
         &          & TinyImageNet (Resize) &                             79.8/81.5/74.1/99.7/\textbf{99.8} &           90.6/90.1/85.2/98.4/\textbf{98.6} &           93.6/93.0/88.9/\textbf{99.9}/99.8 \\
         &          & LSUN (Crop) &                             77.2/79.0/71.5/\textbf{99.2}/98.6 &           89.9/89.4/84.0/\textbf{97.7}/97.0 &           92.9/92.5/86.2/\textbf{99.7}/99.5 \\
         &          & LSUN (Resize) &                             75.5/77.2/67.2/99.5/\textbf{99.9} &           89.1/88.3/82.0/98.3/\textbf{98.9} &  91.6/90.7/84.9/\textbf{99.9}/\textbf{99.9} \\
         &          & iSUN &                             78.1/79.8/69.0/99.8/\textbf{99.9} &           89.9/89.4/82.7/98.7/\textbf{98.9} &  92.3/91.5/85.6/\textbf{99.9}/\textbf{99.9} \\
         &          & Cifar100 &                             77.7/78.9/74.0/\textbf{97.8}/95.5 &           89.5/88.9/85.2/\textbf{96.4}/95.5 &           92.4/91.5/89.1/\textbf{99.3}/98.9 \\
         & densenet & Uniform &                    66.3/64.0/37.9/\textbf{100.}/\textbf{100.} &  87.7/83.6/79.0/\textbf{100.}/\textbf{100.} &  93.2/90.1/87.2/\textbf{100.}/\textbf{100.} \\
         &          & Gaussian &                    88.2/90.8/55.9/\textbf{100.}/\textbf{100.} &  93.5/93.2/84.1/\textbf{100.}/\textbf{100.} &  97.4/97.8/91.2/\textbf{100.}/\textbf{100.} \\
         &          & TinyImageNet (Crop) &                             79.1/77.1/72.4/\textbf{100.}/99.9 &           89.9/87.4/86.6/\textbf{99.4}/99.3 &  94.7/92.8/93.2/\textbf{100.}/\textbf{100.} \\
         &          & TinyImageNet (Resize) &                    80.1/78.7/73.1/\textbf{100.}/\textbf{100.} &           90.3/88.1/87.0/99.5/\textbf{99.6} &  94.9/93.4/93.5/\textbf{100.}/\textbf{100.} \\
         &          & LSUN (Crop) &                             74.0/67.8/71.6/\textbf{99.7}/99.2 &           88.1/83.2/85.5/\textbf{98.7}/97.8 &           93.0/88.5/92.1/\textbf{99.9}/99.8 \\
         &          & LSUN (Resize) &                    77.4/76.8/70.6/\textbf{100.}/\textbf{100.} &  89.3/87.1/85.4/\textbf{99.8}/\textbf{99.8} &  94.2/92.6/92.3/\textbf{100.}/\textbf{100.} \\
         &          & iSUN &                    78.7/78.2/70.2/\textbf{100.}/\textbf{100.} &           89.8/87.6/85.8/99.7/\textbf{99.8} &  94.5/92.9/92.6/\textbf{100.}/\textbf{100.} \\
         &          & Cifar100 &                             68.5/64.6/74.8/\textbf{98.1}/97.9 &           86.6/83.0/87.3/\textbf{96.8}/96.7 &           91.4/88.2/93.9/\textbf{99.5}/99.4 \\
cifar100 & resnet & Uniform &                    12.1/ 1.6/ 1.3/\textbf{100.}/\textbf{100.} &  81.5/84.7/80.6/\textbf{100.}/\textbf{100.} &  85.2/86.5/81.6/\textbf{100.}/\textbf{100.} \\
         &          & Gaussian &                     0.0/ 0.0/ 0.0/\textbf{100.}/\textbf{100.} &  60.7/60.8/81.7/\textbf{100.}/\textbf{100.} &  45.1/46.1/76.6/\textbf{100.}/\textbf{100.} \\
         &          & TinyImageNet (Crop) &                             25.0/27.1/28.5/90.8/\textbf{95.0} &           72.6/74.4/71.2/93.2/\textbf{95.0} &           79.7/81.6/78.9/98.4/\textbf{98.9} \\
         &          & TinyImageNet (Resize) &                             21.3/25.3/25.6/89.4/\textbf{97.9} &           71.0/73.0/68.6/92.6/\textbf{96.7} &           77.2/80.1/76.2/98.1/\textbf{99.5} \\
         &          & LSUN (Crop) &                             19.2/18.0/28.0/\textbf{85.3}/76.2 &           69.6/69.2/77.7/\textbf{91.4}/88.2 &           75.6/75.1/84.3/\textbf{97.5}/95.3 \\
         &          & LSUN (Resize) &                             19.4/23.3/23.0/90.2/\textbf{98.5} &           70.0/71.9/68.5/92.9/\textbf{97.4} &           75.7/78.5/75.5/98.3/\textbf{99.6} \\
         &          & iSUN &                             17.7/20.9/21.5/89.6/\textbf{97.0} &           70.3/72.1/68.5/92.8/\textbf{96.2} &           75.8/78.2/75.3/98.3/\textbf{99.3} \\
         &          & SVHN &                             21.3/18.8/26.3/87.2/\textbf{92.1} &           73.4/73.5/77.3/91.5/\textbf{93.6} &           79.5/79.5/84.2/97.6/\textbf{98.2} \\
         & densenet & Uniform &                     0.0/ 0.0/ 0.0/\textbf{100.}/\textbf{100.} &  64.2/66.7/76.1/\textbf{100.}/\textbf{100.} &  43.3/50.3/74.0/\textbf{100.}/\textbf{100.} \\
         &          & Gaussian &                     0.0/ 0.0/ 0.0/\textbf{100.}/\textbf{100.} &  58.9/53.0/69.8/\textbf{100.}/\textbf{100.} &  30.8/15.6/58.2/\textbf{100.}/\textbf{100.} \\
         &          & TinyImageNet (Crop) &                             23.5/39.3/65.1/\textbf{95.4}/89.2 &           68.8/76.9/86.4/\textbf{95.2}/92.7 &           75.9/84.8/93.7/\textbf{99.1}/97.5 \\
         &          & TinyImageNet (Resize) &                             16.6/22.8/64.4/95.7/\textbf{96.1} &           65.9/71.4/86.8/95.4/\textbf{95.6} &           71.5/78.2/93.8/\textbf{99.2}/98.9 \\
         &          & LSUN (Crop) &                             27.5/51.9/43.8/\textbf{84.9}/68.4 &           72.5/82.2/79.0/\textbf{91.3}/85.0 &           79.8/90.1/87.2/\textbf{97.4}/92.6 \\
         &          & LSUN (Resize) &                             15.8/22.0/72.8/94.3/\textbf{97.9} &           65.0/72.4/88.8/94.8/\textbf{96.7} &           70.6/79.5/95.3/99.0/\textbf{99.3} \\
         &          & iSUN &                             14.2/18.2/68.4/94.0/\textbf{96.3} &           64.0/70.5/86.9/94.6/\textbf{95.7} &           69.5/77.2/93.9/98.9/\textbf{99.0} \\
         &          & SVHN &                             25.2/33.2/33.1/\textbf{88.8}/87.9 &           75.5/80.5/79.5/\textbf{92.3}/91.9 &           82.4/87.7/86.9/\textbf{97.7}/97.1 \\
\bottomrule
\label{reacttable}

\end{tabular}

%%%%%%%%%%%%%%%%%%%%%%%%%%%%%%%%%%%%%%%%%%%%%%%%%%%%%%%%%%%%

}
\end{table}

%% file: feature_selection.tex
\begin{table}[H]
    \centering
    \resizebox{\textwidth}{!}{
    \begin{tabular}{lllllllllllll}
\toprule
{} & \multicolumn{4}{l}{cifar10} & \multicolumn{4}{l}{svhn} & \multicolumn{4}{l}{cifar100} \\
{} & \multicolumn{2}{l}{resnet} & \multicolumn{2}{l}{densenet} & \multicolumn{2}{l}{resnet} & \multicolumn{2}{l}{densenet} & \multicolumn{2}{l}{resnet} & \multicolumn{2}{l}{densenet} \\
{} &         \texttt{XOOD-M} &              \texttt{XOOD-L} &          \texttt{XOOD-M} &              \texttt{XOOD-L} &          \texttt{XOOD-M} &              \texttt{XOOD-L} &          \texttt{XOOD-M} &              \texttt{XOOD-L} &          \texttt{XOOD-M} &              \texttt{XOOD-L} &          \texttt{XOOD-M} &              \texttt{XOOD-L} \\
\midrule
Min \& Max  &  \textbf{99.7} &  \textbf{99.47} &            99.4 &  \textbf{99.55} &  \textbf{99.77} &  \textbf{99.85} &  \textbf{99.94} &  \textbf{99.97} &  \textbf{98.42} &           97.86 &            97.3 &  \textbf{98.29} \\
Min        &          99.63 &            99.4 &           99.23 &           99.41 &           99.43 &           99.77 &           99.87 &           99.93 &           98.17 &  \textbf{98.04} &           96.63 &           96.84 \\
Max        &          99.59 &           99.32 &  \textbf{99.43} &           99.54 &            99.6 &           99.69 &           99.91 &           99.97 &            97.4 &           97.13 &  \textbf{97.37} &           97.48 \\
Positivity &          97.86 &           98.76 &           95.56 &           98.22 &           99.43 &           99.82 &           99.45 &           99.81 &           89.76 &           90.47 &           91.38 &           92.06 \\
Sum        &          97.67 &           98.92 &           96.06 &            97.1 &           99.54 &           99.81 &           99.64 &           99.75 &            92.2 &           96.41 &           92.32 &           92.82 \\
L1         &          98.15 &           98.75 &           98.11 &            98.2 &           98.87 &           99.53 &           99.69 &           99.74 &           93.55 &           95.77 &           92.22 &           94.18 \\
L2         &           98.5 &           98.78 &           98.57 &           98.68 &           98.98 &           99.51 &           99.78 &           99.82 &           93.71 &           96.01 &           93.19 &           94.78 \\
L3         &          98.83 &            98.9 &           98.83 &           98.99 &           99.16 &           99.54 &            99.8 &           99.87 &           94.74 &           95.96 &           94.01 &           95.37 \\
Split L1   &          98.59 &           98.96 &            98.2 &           98.19 &           99.69 &           99.78 &           99.82 &           99.88 &           95.36 &           96.23 &           93.03 &           93.13 \\
Split L2   &          98.94 &           99.08 &           98.61 &           98.89 &           99.61 &           99.72 &           99.86 &           99.89 &           95.94 &           97.36 &            93.6 &           94.74 \\
Split L3   &          99.23 &           99.18 &           98.92 &           99.24 &           99.69 &           99.82 &           99.88 &           99.92 &            96.8 &            97.5 &           94.33 &           95.66 \\
\bottomrule
\end{tabular}
}
    \caption{AUROC-score when replacing the extreme values with other features which have linear computational complexity.}
    \label{Table: Feature Selection}
\end{table}

%% file: feature_addition.tex
\begin{table}[H]
    \centering
    \resizebox{\textwidth}{!}{
    \begin{tabular}{lllllllllllll}
\toprule
{} & \multicolumn{4}{l}{cifar10} & \multicolumn{4}{l}{svhn} & \multicolumn{4}{l}{cifar100} \\
{} & \multicolumn{2}{l}{resnet} & \multicolumn{2}{l}{densenet} & \multicolumn{2}{l}{resnet} & \multicolumn{2}{l}{densenet} & \multicolumn{2}{l}{resnet} & \multicolumn{2}{l}{densenet} \\
{} &         \texttt{XOOD-M} &              \texttt{XOOD-L} &         \texttt{XOOD-M} &             \texttt{XOOD-L} &         \texttt{XOOD-M} &              \texttt{XOOD-L} &          \texttt{XOOD-M} &              \texttt{XOOD-L} &          \texttt{XOOD-M} &              \texttt{XOOD-L} &         \texttt{XOOD-M} &              \texttt{XOOD-L} \\
\midrule
Min \& Max  &  \textbf{99.7} &           99.46 &  \textbf{99.4} &  \textbf{99.5} &          99.77 &           99.85 &           99.94 &           99.97 &  \textbf{98.42} &            97.8 &  \textbf{97.3} &  \textbf{98.18} \\
Positivity &          99.68 &  \textbf{99.54} &          99.27 &          99.34 &          99.88 &           99.92 &  \textbf{99.95} &  \textbf{99.98} &           98.14 &           97.21 &           97.0 &           97.52 \\
Sum        &          99.64 &            99.5 &          99.28 &          99.19 &  \textbf{99.9} &  \textbf{99.93} &           99.94 &           99.96 &           98.22 &           97.66 &          96.85 &           97.45 \\
L1         &          99.62 &           99.43 &          99.33 &          99.38 &          99.79 &           99.87 &           99.93 &           99.92 &            98.3 &            97.5 &          96.63 &           97.74 \\
L2         &          99.63 &           99.42 &          99.37 &          99.39 &          99.79 &           99.78 &           99.94 &           99.92 &           98.18 &           97.61 &          96.69 &           97.86 \\
L3         &          99.63 &           99.43 &          99.39 &          99.39 &           99.8 &           99.78 &           99.94 &           99.93 &           98.18 &           97.58 &          96.76 &           97.97 \\
Split L1   &          99.57 &           99.43 &          99.26 &          99.17 &           99.9 &           99.91 &           99.93 &           99.95 &           98.21 &            97.5 &          96.33 &           97.11 \\
Split L2   &          99.59 &           99.41 &          99.29 &          99.22 &          99.86 &           99.87 &           99.94 &           99.94 &           98.13 &           97.82 &          96.28 &           97.51 \\
Split L3   &          99.62 &           99.44 &          99.33 &          99.25 &          99.87 &           99.87 &           99.93 &           99.94 &           98.21 &  \textbf{97.85} &          96.36 &           97.74 \\
\bottomrule
\end{tabular}
}
    \caption{AUROC-score when extracting additional extreme values before each activation layer.}
    \label{Table: Feature Addition}
\end{table}

%% file: fpr_table.tex
\begin{table}[H]
    \centering
    \begin{tabular}{lrrrrrr}
        & Baseline & ODIN & Mahalanobis & Gram & XOOD-L & XOOD-M \\
        \hline
        ResNet34 & 55.7	& 37.6 & 8.6& 7.1& 5.2 & 3.5 \\
        DenseNet & 51.8	& 28.4 & 15.2 & 6.9 & 3.6 & 5.1\\
    \end{tabular}
    \caption{Average FPR at 95\% TPR. Lower score is better.}
    \label{fpr-table}
\end{table}